\documentclass{article}

\usepackage[final]{neurips_2019}

\usepackage[utf8]{inputenc} 
\usepackage[T1]{fontenc}    
\usepackage{hyperref}       
\usepackage{url}            
\usepackage{booktabs}       
\usepackage{amsfonts}       
\usepackage{nicefrac}       
\usepackage{microtype}      
\usepackage[pdftex]{graphicx}
\usepackage{wrapfig}

\title{DistilBERT, a distilled version of BERT: smaller, faster, cheaper and lighter}

\author{%
  Victor SANH, Lysandre DEBUT, Julien CHAUMOND, Thomas WOLF\\
  Hugging Face\\
  \texttt{\{victor,lysandre,julien,thomas\}@huggingface.co}
}

\begin{document}

\maketitle

\begin{abstract}
    As Transfer Learning from large-scale pre-trained models becomes more prevalent in Natural Language Processing (NLP), operating these large models in on-the-edge and/or under constrained computational training or inference budgets remains challenging. In this work, we propose a method to pre-train a smaller general-purpose language representation model, called DistilBERT, which can then be fine-tuned with good performances on a wide range of tasks like its larger counterparts. While most prior work investigated the use of distillation for building task-specific models, we leverage knowledge distillation during the pre-training phase and show that it is possible to reduce the size of a BERT model by 40\%, while retaining 97\% of its language understanding capabilities and being 60\% faster. To leverage the inductive biases learned by larger models during pre-training, we introduce a triple loss combining language modeling, distillation and cosine-distance losses. Our smaller, faster and lighter model is cheaper to pre-train and we demonstrate its capabilities for on-device computations in a proof-of-concept experiment and a comparative on-device study.
\end{abstract}

\section{Introduction}

\begin{wrapfigure}{r}{0.61\textwidth}
  \vspace{-15pt}
  \begin{center}
    \includegraphics[width=\linewidth]{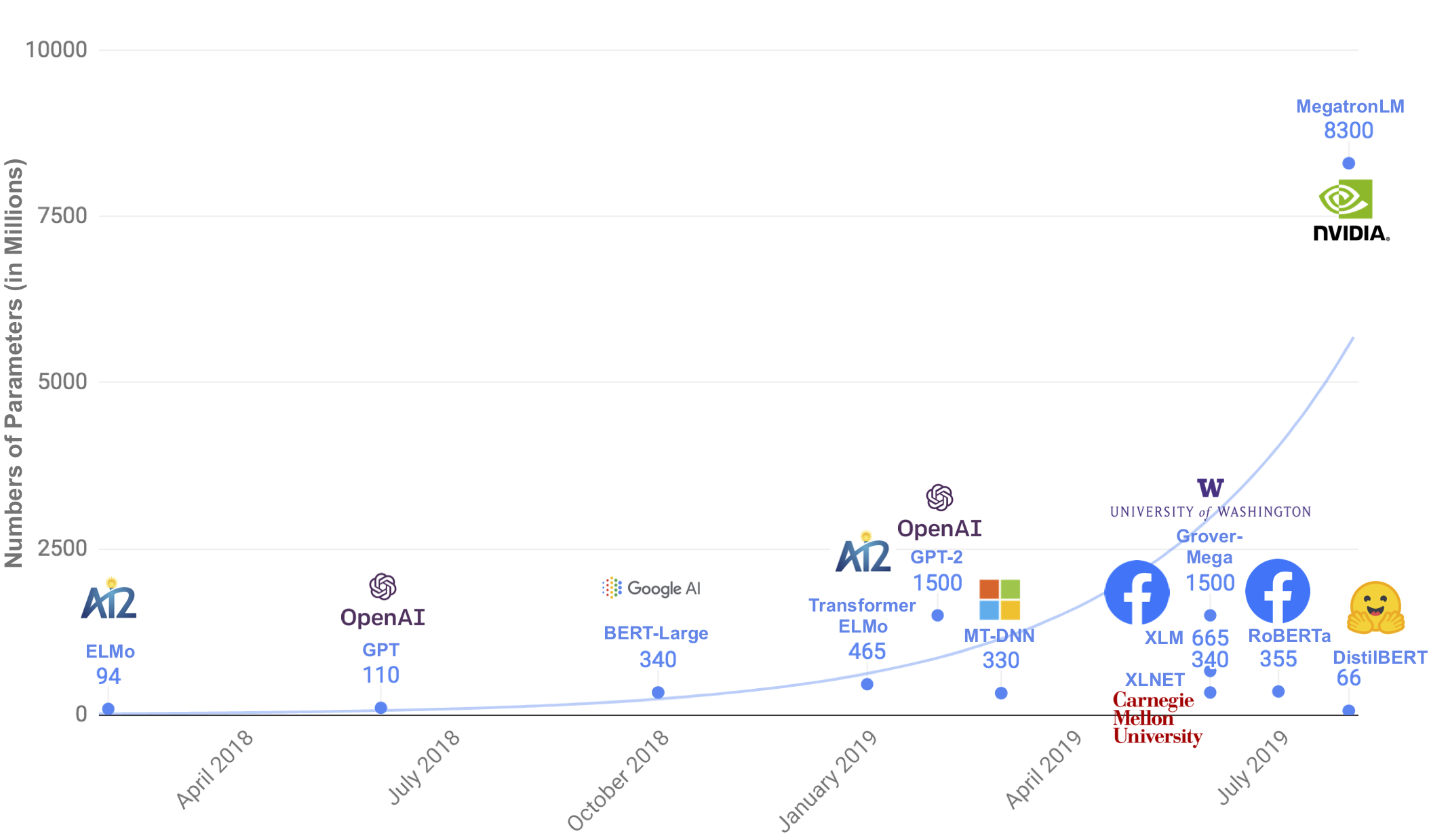}
  \end{center}
  \vspace{-10pt}
  \caption{\textbf{Parameter counts of several recently released pretrained language models.}}
  \vspace{-10pt}
\end{wrapfigure}

The last two years have seen the rise of Transfer Learning approaches in Natural Language Processing (NLP) with large-scale pre-trained language models becoming a basic tool in many NLP tasks \citep{Devlin2018BERTPO, Radford2019LanguageMA, Liu2019RoBERTaAR}. While these models lead to significant improvement, they often have several hundred million parameters and current research\footnote{See for instance the recently released MegatronLM (\url{https://nv-adlr.github.io/MegatronLM})} on pre-trained models indicates that training even larger models still leads to better performances on downstream tasks.

The trend toward bigger models raises several concerns. First is the environmental cost of exponentially scaling these models' computational requirements as mentioned in \citet{Schwartz2019GreenA, Strubell2019EnergyAP}. Second, while operating these models on-device in real-time has the potential to enable novel and interesting language processing applications, the growing computational and memory requirements of these models may hamper wide adoption.

In this paper, we show that it is possible to reach similar performances on many downstream-tasks using much smaller language models pre-trained with knowledge distillation, resulting in models that are lighter and faster at inference time, while also requiring a smaller computational training budget. Our general-purpose pre-trained models can be fine-tuned with good performances on several downstream tasks, keeping the flexibility of larger models. We also show that our compressed models are small enough to run on the edge, e.g. on mobile devices.

Using a triple loss, we show that a 40\% smaller Transformer (\citet{Vaswani2017AttentionIA}) pre-trained through distillation via the supervision of a bigger Transformer language model can achieve similar performance on a variety of downstream tasks, while being 60\% faster at inference time.  Further ablation studies indicate that all the components of the triple loss are important for best performances.

We have made the trained weights available along with the training code in the \texttt{Transformers}\footnote{https://github.com/huggingface/transformers} library from HuggingFace~\citep{wolf2019transformers}.

\section{Knowledge distillation}

\textbf{Knowledge distillation} \citep{Bucila2006ModelC, Hinton2015DistillingTK} is a compression technique in which a compact model - the student - is trained to reproduce the behaviour of a larger model - the teacher - or an ensemble of models.

In supervised learning, a classification model is generally trained to predict an instance class by maximizing the estimated probability of gold labels. A standard training objective thus involves minimizing the cross-entropy between the model's predicted distribution and the one-hot empirical distribution of training labels. A model performing well on the training set will predict an output distribution with high probability on the correct class and with near-zero probabilities on other classes. But some of these "near-zero" probabilities are larger than others and reflect, in part, the generalization capabilities of the model and how well it will perform on the test set\footnote{E.g. BERT-base's predictions for a masked token in "\texttt{I think this is the beginning of a beautiful \lbrack MASK\rbrack}" comprise two high probability tokens (\textit{day} and \textit{life}) and a long tail of valid predictions (\textit{future}, \textit{story}, \textit{world}\dots).
}.

\textbf{Training loss} The student is trained with a distillation loss over the soft target probabilities of the teacher: $L_{ce} = \sum_i t_i * \log (s_i)$ where $t_i$ (resp. $s_i$) is a probability estimated by the teacher (resp. the student). This objective results in a rich training signal by leveraging the full teacher distribution. Following \citet{Hinton2015DistillingTK} we used a \textit{softmax-temperature}: $p_i = \frac{\exp (z_i / T)}{\sum_j \exp(z_j / T)}$ where $T$ controls the smoothness of the output distribution and $z_i$ is the model score for the class $i$. The same temperature $T$ is applied to the student and the teacher at training time, while at inference, $T$ is set to 1 to recover a standard \textit{softmax}.

The final training objective is a linear combination of the distillation loss $L_{ce}$ with the supervised training loss, in our case the \textit{masked language modeling} loss $L_{mlm}$ \citep{Devlin2018BERTPO}. We found it beneficial to add a \textit{cosine embedding} loss ($L_{cos}$) which will tend to align the directions of the student and teacher hidden states vectors.

\section{DistilBERT: a distilled version of BERT}

\textbf{Student architecture} In the present work, the student - DistilBERT - has the same general architecture as BERT. The \textit{token-type embeddings} and the \textit{pooler} are removed while the number of layers is reduced by a factor of 2. Most of the operations used in the Transformer architecture (\textit{linear layer} and \textit{layer normalisation}) are highly optimized in modern linear algebra frameworks and our investigations showed that variations on the last dimension of the tensor (hidden size dimension) have a smaller impact on computation efficiency (for a fixed parameters budget) than variations on other factors like the number of layers. Thus we focus on reducing the number of layers.

\textbf{Student initialization} In addition to the previously described optimization and architectural choices, an important element in our training procedure is to find the right initialization for the sub-network to converge. Taking advantage of the common dimensionality between teacher and student networks, we initialize the student from the teacher by taking one layer out of two.

\textbf{Distillation} We applied best practices for training BERT model recently proposed in \citet{Liu2019RoBERTaAR}. As such, DistilBERT is distilled on very large batches leveraging gradient accumulation (up to 4K examples per batch) using dynamic masking and without the next sentence prediction objective.

\textbf{Data and compute power} We train DistilBERT on the same corpus as the original BERT model: a concatenation of English Wikipedia and \texttt{Toronto Book Corpus} \citep{Zhu2015AligningBA}. DistilBERT was trained on 8 16GB V100 GPUs for approximately 90 hours. For the sake of comparison, the RoBERTa model \citep{Liu2019RoBERTaAR} required 1 day of training on 1024 32GB V100.

\section{Experiments}

\textbf{General Language Understanding} We assess the language understanding and generalization capabilities of DistilBERT on the \textit{General Language Understanding Evaluation} (GLUE) benchmark \citep{Wang2018GLUEAM}, a collection of 9 datasets for evaluating natural language understanding systems. We report scores on the development sets for each task by fine-tuning DistilBERT without the use of ensembling or multi-tasking scheme for fine-tuning (which are mostly orthogonal to the present work). We compare the results to the baseline provided by the authors of GLUE: an ELMo (\citet{Peters2018DeepCW}) encoder followed by two BiLSTMs.\footnote{We use \texttt{jiant} \citep{wang2019jiant} to compute the baseline.}

The results on each of the 9 tasks are showed on Table~\ref{glue_benchmark} along with the macro-score (average of individual scores). Among the 9 tasks, DistilBERT is always on par or improving over the ELMo baseline (up to 19 points of accuracy on STS-B). DistilBERT also compares surprisingly well to BERT, retaining 97\% of the performance with 40\% fewer parameters.

\begin{table}
  \caption{\textbf{DistilBERT retains 97\% of BERT performance.} Comparison on the dev sets of the GLUE benchmark. ELMo results as reported by the authors. BERT and DistilBERT results are the medians of 5 runs with different seeds.}
  \label{glue_benchmark}
  \centering
  \resizebox{\columnwidth}{!}{%
    \begin{tabular}{lcccccccccc}
        \toprule
        Model & \textbf{Score} & CoLA & MNLI & MRPC & QNLI & QQP & RTE & SST-2 & STS-B & WNLI\\
        \cmidrule(r){2-11}
        ELMo & 68.7 & 44.1 & 68.6 & 76.6 & 71.1 & 86.2 & 53.4 & 91.5 & 70.4 & 56.3 \\
        BERT-base & 79.5 & 56.3 & 86.7 & 88.6 & 91.8 & 89.6 & 69.3 & 92.7 & 89.0 & 53.5 \\
        DistilBERT & 77.0 & 51.3 & 82.2 & 87.5 & 89.2 & 88.5 & 59.9 & 91.3 & 86.9 & 56.3 \\
        \bottomrule
    \end{tabular}%
    }
\end{table}

\subsection{Downstream task benchmark}

\textbf{Downstream tasks} We further study the performances of DistilBERT on several downstream tasks under efficient inference constraints: a classification task (IMDb sentiment classification - \citet{Maas2011LearningWV}) and a question answering task (SQuAD v1.1 - \citet{Rajpurkar2016SQuAD10}).

\begin{table}
\parbox{.45\linewidth}{
\centering
  \caption{\textbf{DistilBERT yields to comparable performance on downstream tasks.} Comparison on downstream tasks: IMDb (test accuracy) and SQuAD 1.1 (EM/F1 on dev set). D: with a second step of distillation during fine-tuning.}
  \label{downstream}
  \centering
  \begin{tabular}{lcc}
    \toprule
    Model & IMDb & SQuAD \\
    & (acc.) & (EM/F1) \\
    \cmidrule(r){2-3}
    BERT-base & 93.46 & 81.2/88.5 \\
    DistilBERT & 92.82 & 77.7/85.8 \\
    DistilBERT (D) & - & 79.1/86.9 \\
    \bottomrule
  \end{tabular}
}
\hfill
\parbox{.45\linewidth}{
  \caption{\textbf{DistilBERT is significantly smaller while being constantly faster.} Inference time of a full pass of GLUE task STS-B (sentiment analysis) on CPU with a batch size of 1.}
  \label{inference}
  \centering
  \begin{tabular}{lcc}
    \toprule
    Model & \# param. & Inf. time \\
     & (Millions) & (seconds) \\
    \cmidrule(r){2-3}
    ELMo & 180 & 895 \\
    BERT-base & 110 & 668 \\
    DistilBERT & 66 & 410 \\
    \bottomrule
  \end{tabular}
}
\end{table}

As shown in Table~\ref{downstream}, DistilBERT is only 0.6\% point behind BERT in test accuracy on the IMDb benchmark while being 40\% smaller. On SQuAD, DistilBERT is within 3.9 points of the full BERT.

We also studied whether we could add another step of distillation during the adaptation phase by fine-tuning DistilBERT on SQuAD using a BERT model previously fine-tuned on SQuAD as a teacher for an additional term in the loss (knowledge distillation). In this setting, there are thus two successive steps of distillation, one during the pre-training phase and one during the adaptation phase. In this case, we were able to reach interesting performances given the size of the model: 79.8 F1 and 70.4 EM, i.e. within 3 points of the full model.

\textbf{Size and inference speed}

To further investigate the speed-up/size trade-off of DistilBERT, we compare (in Table~\ref{inference}) the number of parameters of each model along with the inference time needed to do a full pass on the STS-B development set on CPU (Intel Xeon E5-2690 v3 Haswell @2.9GHz) using a batch size of 1. DistilBERT has 40\% fewer parameters than BERT and is 60\% faster than BERT.

\textbf{On device computation}
We studied whether DistilBERT could be used for on-the-edge applications by building a mobile application for question answering. We compare the average inference time on a recent smartphone (iPhone 7 Plus) against our previously trained question answering model based on BERT-base. Excluding the tokenization step, DistilBERT is 71\% faster than BERT, and the whole model weighs 207 MB (which could be further reduced with quantization). Our code is available\footnote{https://github.com/huggingface/swift-coreml-transformers}.

\subsection{Ablation study}

In this section, we investigate the influence of various components of the triple loss and the student initialization on the performances of the distilled model. We report the macro-score on GLUE. Table~\ref{ablation} presents the deltas with the full triple loss: removing the \textit{Masked Language Modeling} loss has little impact while the two distillation losses account for a large portion of the performance.

\begin{table}
  \caption{\textbf{Ablation study.} Variations are relative to the model trained with triple loss and teacher weights initialization.}
  \label{ablation}
  \centering
  \begin{tabular}{lc}
    \toprule
    Ablation & \textbf{Variation on GLUE macro-score}\\
    \cmidrule(r){1-2}
    $\emptyset$ - $L_{cos}$ - $L_{mlm}$ & -2.96\\
    $L_{ce}$ - $\emptyset$ - $L_{mlm}$ & -1.46\\
    $L_{ce}$ - $L_{cos}$ - $\emptyset$ & -0.31\\
    Triple loss + random weights initialization & -3.69\\
    \bottomrule
  \end{tabular}
\end{table}

\section{Related work}

\textbf{Task-specific distillation} Most of the prior works focus on building task-specific distillation setups. \citet{Tang2019DistillingTK} transfer fine-tune classification model BERT to an LSTM-based classifier. \citet{Chatterjee2019MakingNM} distill BERT model fine-tuned on SQuAD in a smaller Transformer model previously initialized from BERT. In the present work, we found it beneficial to use a general-purpose pre-training distillation rather than a task-specific distillation. \citet{Turc2019WellReadSL} use the original pretraining objective to train smaller student, then fine-tuned via distillation. As shown in the ablation study, we found it beneficial to leverage the teacher's knowledge to pre-train with additional distillation signal.

\textbf{Multi-distillation} \citet{Yang2019ModelCW} combine the knowledge of an ensemble of teachers using multi-task learning to regularize the distillation. The authors apply \textit{Multi-Task Knowledge Distillation} to learn a compact question answering model from a set of large question answering models. An application of multi-distillation is multi-linguality: \citet{Tsai2019SmallAP} adopts a similar approach to us by pre-training a multilingual model from scratch solely through distillation. However, as shown in the ablation study, leveraging the teacher's knowledge with initialization and additional losses leads to substantial gains.

\textbf{Other compression techniques} have been studied to compress large models. Recent developments in weights pruning reveal that it is possible to remove some heads in the self-attention at test time without significantly degrading the performance \citet{Michel2019AreSH}. Some layers can be reduced to one head. A separate line of study leverages quantization to derive smaller models (\citet{Gupta2015DeepLW}). Pruning and quantization are orthogonal to the present work.

\section{Conclusion and future work}

We introduced DistilBERT, a general-purpose pre-trained version of BERT, 40\% smaller, 60\% faster, that retains 97\% of the language understanding capabilities. We showed that a general-purpose language model can be successfully trained with distillation and analyzed the various components with an ablation study. We further demonstrated that DistilBERT is a compelling option for edge applications.

\fontsize{9.0pt}{9.0pt} \selectfont
\bibliography{neurips_2019} 

\begin{thebibliography}{22}
\providecommand{\natexlab}[1]{#1}
\providecommand{\url}[1]{\texttt{#1}}
\expandafter\ifx\csname urlstyle\endcsname\relax
  \providecommand{\doi}[1]{doi: #1}\else
  \providecommand{\doi}{doi: \begingroup \urlstyle{rm}\Url}\fi

\bibitem[Devlin et~al.(2018)Devlin, Chang, Lee, and
  Toutanova]{Devlin2018BERTPO}
Jacob Devlin, Ming-Wei Chang, Kenton Lee, and Kristina Toutanova.
\newblock Bert: Pre-training of deep bidirectional transformers for language
  understanding.
\newblock In \emph{NAACL-HLT}, 2018.

\bibitem[Radford et~al.(2019)Radford, Wu, Child, Luan, Amodei, and
  Sutskever]{Radford2019LanguageMA}
Alec Radford, Jeffrey Wu, Rewon Child, David Luan, Dario Amodei, and Ilya
  Sutskever.
\newblock Language models are unsupervised multitask learners.
\newblock 2019.

\bibitem[Liu et~al.(2019)Liu, Ott, Goyal, Du, Joshi, Chen, Levy, Lewis,
  Zettlemoyer, and Stoyanov]{Liu2019RoBERTaAR}
Yinhan Liu, Myle Ott, Naman Goyal, Jingfei Du, Mandar~S. Joshi, Danqi Chen,
  Omer Levy, Mike Lewis, Luke~S. Zettlemoyer, and Veselin Stoyanov.
\newblock Roberta: A robustly optimized bert pretraining approach.
\newblock \emph{ArXiv}, abs/1907.11692, 2019.

\bibitem[Schwartz et~al.(2019)Schwartz, Dodge, Smith, and
  Etzioni]{Schwartz2019GreenA}
Roy Schwartz, Jesse Dodge, Noah~A. Smith, and Oren Etzioni.
\newblock Green ai.
\newblock \emph{ArXiv}, abs/1907.10597, 2019.

\bibitem[Strubell et~al.(2019)Strubell, Ganesh, and
  McCallum]{Strubell2019EnergyAP}
Emma Strubell, Ananya Ganesh, and Andrew McCallum.
\newblock Energy and policy considerations for deep learning in nlp.
\newblock In \emph{ACL}, 2019.

\bibitem[Vaswani et~al.(2017)Vaswani, Shazeer, Parmar, Uszkoreit, Jones, Gomez,
  Kaiser, and Polosukhin]{Vaswani2017AttentionIA}
Ashish Vaswani, Noam Shazeer, Niki Parmar, Jakob Uszkoreit, Llion Jones,
  Aidan~N. Gomez, Lukasz Kaiser, and Illia Polosukhin.
\newblock Attention is all you need.
\newblock In \emph{NIPS}, 2017.

\bibitem[Wolf et~al.(2019)Wolf, Debut, Sanh, Chaumond, Delangue, Moi, Cistac,
  Rault, Louf, Funtowicz, and Brew]{wolf2019transformers}
Thomas Wolf, Lysandre Debut, Victor Sanh, Julien Chaumond, Clement Delangue,
  Anthony Moi, Pierric Cistac, Tim Rault, Rémi Louf, Morgan Funtowicz, and
  Jamie Brew.
\newblock Transformers: State-of-the-art natural language processing, 2019.

\bibitem[Bucila et~al.(2006)Bucila, Caruana, and
  Niculescu-Mizil]{Bucila2006ModelC}
Cristian Bucila, Rich Caruana, and Alexandru Niculescu-Mizil.
\newblock Model compression.
\newblock In \emph{KDD}, 2006.

\bibitem[Hinton et~al.(2015)Hinton, Vinyals, and Dean]{Hinton2015DistillingTK}
Geoffrey~E. Hinton, Oriol Vinyals, and Jeffrey Dean.
\newblock Distilling the knowledge in a neural network.
\newblock \emph{ArXiv}, abs/1503.02531, 2015.

\bibitem[Zhu et~al.(2015)Zhu, Kiros, Zemel, Salakhutdinov, Urtasun, Torralba,
  and Fidler]{Zhu2015AligningBA}
Yukun Zhu, Ryan Kiros, Richard~S. Zemel, Ruslan Salakhutdinov, Raquel Urtasun,
  Antonio Torralba, and Sanja Fidler.
\newblock Aligning books and movies: Towards story-like visual explanations by
  watching movies and reading books.
\newblock \emph{2015 IEEE International Conference on Computer Vision (ICCV)},
  pages 19--27, 2015.

\bibitem[Wang et~al.(2018)Wang, Singh, Michael, Hill, Levy, and
  Bowman]{Wang2018GLUEAM}
Alex Wang, Amanpreet Singh, Julian Michael, Felix Hill, Omer Levy, and
  Samuel~R. Bowman.
\newblock Glue: A multi-task benchmark and analysis platform for natural
  language understanding.
\newblock In \emph{ICLR}, 2018.

\bibitem[Peters et~al.(2018)Peters, Neumann, Iyyer, Gardner, Clark, Lee, and
  Zettlemoyer]{Peters2018DeepCW}
Matthew~E. Peters, Mark Neumann, Mohit Iyyer, Matt Gardner, Christopher Clark,
  Kenton Lee, and Luke Zettlemoyer.
\newblock Deep contextualized word representations.
\newblock In \emph{NAACL}, 2018.

\bibitem[Wang et~al.(2019)Wang, Tenney, Pruksachatkun, Yu, Hula, Xia,
  Pappagari, Jin, McCoy, Patel, Huang, Phang, Grave, Kim, Htut, F'{e}vry, Chen,
  Nangia, Liu, Mohananey, Bordia, Patry, Pavlick, and Bowman]{wang2019jiant}
Alex Wang, Ian~F. Tenney, Yada Pruksachatkun, Katherin Yu, Jan Hula, Patrick
  Xia, Raghu Pappagari, Shuning Jin, R.~Thomas McCoy, Roma Patel, Yinghui
  Huang, Jason Phang, Edouard Grave, Najoung Kim, Phu~Mon Htut, Thibault
  F'{e}vry, Berlin Chen, Nikita Nangia, Haokun Liu, Anhad Mohananey, Shikha
  Bordia, Nicolas Patry, Ellie Pavlick, and Samuel~R. Bowman.
\newblock \texttt{jiant} 1.1: A software toolkit for research on
  general-purpose text understanding models.
\newblock \url{http://jiant.info/}, 2019.

\bibitem[Maas et~al.(2011)Maas, Daly, Pham, Huang, Ng, and
  Potts]{Maas2011LearningWV}
Andrew~L. Maas, Raymond~E. Daly, Peter~T. Pham, Dan Huang, Andrew~Y. Ng, and
  Christopher Potts.
\newblock Learning word vectors for sentiment analysis.
\newblock In \emph{ACL}, 2011.

\bibitem[Rajpurkar et~al.(2016)Rajpurkar, Zhang, Lopyrev, and
  Liang]{Rajpurkar2016SQuAD10}
Pranav Rajpurkar, Jian Zhang, Konstantin Lopyrev, and Percy Liang.
\newblock Squad: 100, 000+ questions for machine comprehension of text.
\newblock In \emph{EMNLP}, 2016.

\bibitem[Tang et~al.(2019)Tang, Lu, Liu, Mou, Vechtomova, and
  Lin]{Tang2019DistillingTK}
Raphael Tang, Yao Lu, Linqing Liu, Lili Mou, Olga Vechtomova, and Jimmy Lin.
\newblock Distilling task-specific knowledge from bert into simple neural
  networks.
\newblock \emph{ArXiv}, abs/1903.12136, 2019.

\bibitem[Chatterjee(2019)]{Chatterjee2019MakingNM}
Debajyoti Chatterjee.
\newblock Making neural machine reading comprehension faster.
\newblock \emph{ArXiv}, abs/1904.00796, 2019.

\bibitem[Turc et~al.(2019)Turc, Chang, Lee, and Toutanova]{Turc2019WellReadSL}
Iulia Turc, Ming-Wei Chang, Kenton Lee, and Kristina Toutanova.
\newblock Well-read students learn better: The impact of student initialization
  on knowledge distillation.
\newblock \emph{ArXiv}, abs/1908.08962, 2019.

\bibitem[Yang et~al.(2019)Yang, Shou, Gong, Lin, and Jiang]{Yang2019ModelCW}
Ze~Yang, Linjun Shou, Ming Gong, Wutao Lin, and Daxin Jiang.
\newblock Model compression with multi-task knowledge distillation for
  web-scale question answering system.
\newblock \emph{ArXiv}, abs/1904.09636, 2019.

\bibitem[Tsai et~al.(2019)Tsai, Riesa, Johnson, Arivazhagan, Li, and
  Archer]{Tsai2019SmallAP}
Henry Tsai, Jason Riesa, Melvin Johnson, Naveen Arivazhagan, Xin Li, and Amelia
  Archer.
\newblock Small and practical bert models for sequence labeling.
\newblock In \emph{EMNLP-IJCNLP}, 2019.

\bibitem[Michel et~al.(2019)Michel, Levy, and Neubig]{Michel2019AreSH}
Paul Michel, Omer Levy, and Graham Neubig.
\newblock Are sixteen heads really better than one?
\newblock In \emph{NeurIPS}, 2019.

\bibitem[Gupta et~al.(2015)Gupta, Agrawal, Gopalakrishnan, and
  Narayanan]{Gupta2015DeepLW}
Suyog Gupta, Ankur Agrawal, Kailash Gopalakrishnan, and Pritish Narayanan.
\newblock Deep learning with limited numerical precision.
\newblock In \emph{ICML}, 2015.

\end{thebibliography}
\bibliographystyle{unsrtnat}

\end{document}